\pdfoutput=1

\documentclass[11pt]{article}

\usepackage[]{acl}

\usepackage{times}
\usepackage{latexsym}
\usepackage{stfloats}
\usepackage[bottom]{footmisc}
\usepackage{tcolorbox}

\usepackage[T1]{fontenc}

\usepackage[utf8]{inputenc}

\usepackage{graphicx}
\usepackage{booktabs}
\usepackage{adjustbox}
\usepackage{amsmath}
\usepackage{times}  
\usepackage{helvet}  
\usepackage{courier}  
\usepackage{algorithm}
\usepackage[noend]{algpseudocode}
\usepackage{tensor}
\usepackage{amsfonts}
\usepackage{float}
\usepackage{ragged2e}
\usepackage{tabularx}
\usepackage{hyperref}

\title{SparseGrad: A Selective Method for Efficient Fine-tuning of MLP Layers}

\author{
  \textbf{Viktoriia Chekalina\textsuperscript{\rm 1,2}}\quad
  \textbf{Anna Rudenko\textsuperscript{\rm 1,2}} \quad
  \textbf{Gleb Mezentsev\textsuperscript{\rm 1,2}} \quad \\
  \textbf{Alexander Mikhalev\textsuperscript{\rm 2}} \quad 
  \textbf{Alexander Panchenko\textsuperscript{\rm 2,1}} \quad 
  \textbf{Ivan Oseledets\textsuperscript{\rm 1,2}} \\
  \textsuperscript{\rm 1}Artificial Intelligence Research Institute, \\
  \textsuperscript{\rm 2}Skolkovo Institute of Science and Technology}

\begin{document}
\maketitle
\begin{abstract}
The performance of Transformer models has been enhanced by increasing the number of parameters and the length of the processed text. Consequently, fine-tuning the entire model becomes a memory-intensive process. High-performance methods for parameter-efficient fine-tuning (PEFT) typically work with Attention blocks and often overlook MLP blocks, which contain about half of the model parameters. We propose a new selective PEFT method, namely SparseGrad, that performs well on MLP blocks. We transfer layer gradients to a space where only about 1\% of the layer's elements remain significant. By converting gradients into a sparse structure, we reduce the number of updated parameters. We apply SparseGrad to fine-tune BERT and RoBERTa for the NLU task and LLaMa-2 for the Question-Answering task. In these experiments, with identical memory requirements, our method outperforms LoRA and MeProp, robust popular state-of-the-art PEFT approaches.
\end{abstract}


\section{Introduction}



Due to the tendency to increase the size of transformer models with each new generation, we need efficient ways to fine-tune such models on downstream task data. The usual practice is fine-tuning a large pre-trained foundational model on a downstream task. The major problem that prevents efficient fine-tuning is a steady increase in the memory footprint. One of the best strategies is high-performance methods for parameter-efficient fine-tuning~(PEFT). Typically, such methods as LoRA~\cite{LORA} focus on attention blocks and do not consider dense MLP blocks. Since MLP blocks can take a significant fraction of the model parameters (see Table~\ref{Tab::parameters_per_layers}), we propose to focus instead on MLP blocks. We introduce a novel selective PEFT approach called SparseGrad. Our method is based on finding a special sparsification transformation that allows us to fine-tune about $1\%$  of the dense MLP layer parameters and still show good performance in downstream tasks.

\begin{table}[h]
\caption{Number of parameters for different layers in models based on the Transformer.}
\centering
\scalebox{0.8}{
\footnotesize
\begin{tabular}{l | c |c|c | c |c|c }
\toprule
Blocks/Model & \multicolumn{2}{c|}{BERT} &\multicolumn{2}{c|}{RoBERTa\textsubscript{base}} &\multicolumn{2}{c}{LLaMa-2} \\ 
\midrule
Full model & 109 M & 100\% & 125 M & 100\% &  6.7 B & 100\% \\
MLP & 57 M & 52\% & 57 M  & 45\%  & 4.3 B & 64\%\\
Embeddings & 24 M& 22\%   & 40 M & 32\% &  0.1 B & 1\% \\
Attention & 28 M & 25\% & 28 M & 22\% & 2.1 B & 31\%\\
\bottomrule
\end{tabular}
}
\label{Tab::parameters_per_layers}
\end{table}

We validate our approach on BERT~\cite{devlin2018bert} and RoBERTa~\cite{liu2019roberta} models on GLUE~\cite{wang2018glue} benchmark and in both cases obtain results better than LoRA~\cite{LORA} and MeProp~\cite{meProp} methods. We also fine-tune LLaMa-2~\cite{touvron2023llama} 2.7B on the OpenAssistant dataset~\cite{kopf2023openassistant} and also achieve performance higher than LoRA and MeProp.

\section{Related Work}
In the last few years, many approaches to PEFT have appeared.
\citet{lialin2023scaling} distinguishes three types of methods: additive, reparametrization-based, and selective. In additive PEFT, small neural networks called adapters are added to the main model to steer the outputs of its modules~\cite{adapter_hub}. Adapters are trainable, therefore, the main model remains unchanged.~\citet{pmlr-v97-houlsby19a} adapt this approach to NLP. In reparametrization-based approaches low-rank representations of trainable parameters are used. For example, LoRA~\cite{LORA} parameterizes the weight update by a trainable low-rank matrix decomposition. In the original paper, LoRA is applied to self-attention modules, but not to MLP ones. In the selective methods, parts of the model or sets of the parameters are chosen for fine-tuning using some heuristics. Such methods include, for example, Bit Fit~\cite{DBLP:journals/corr/abs-2106-10199} or MeProp~\cite{meProp}, where only top-k parameters are updated during backpropagation. The approach proposed in this paper is related to selective methods. 

\begin{figure*}[ht]
\centering
\scalebox{0.9}{
\includegraphics[width=0.9\linewidth]{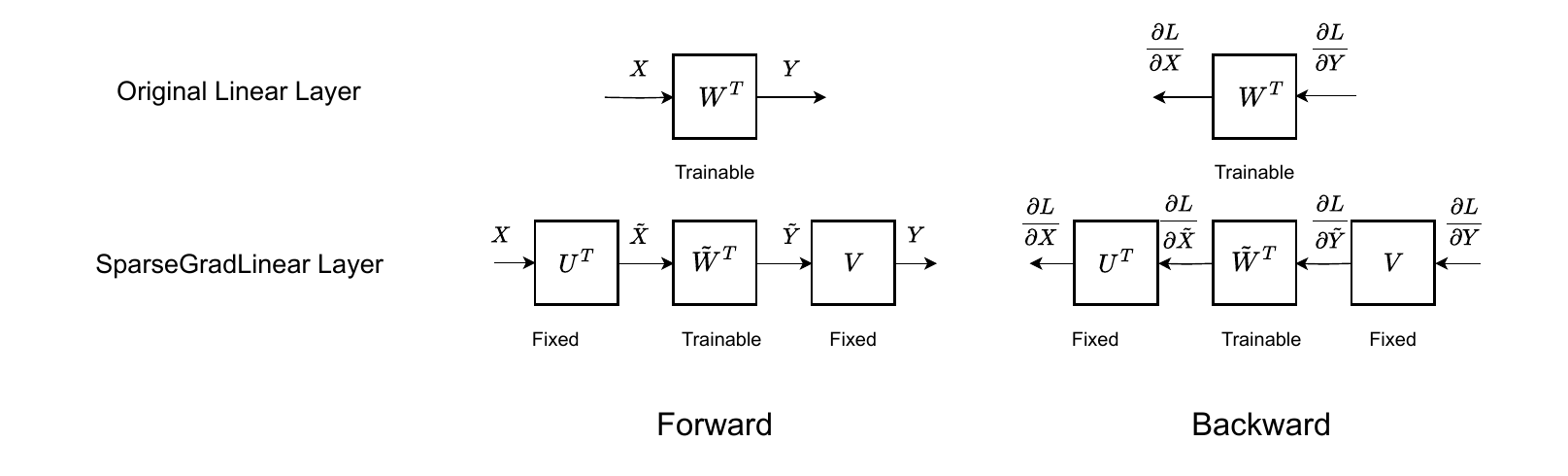}
}
\caption{The first row illustrates signal propagation in the original Linear Layer, while the second row illustrates propagation with the proposed SparseGradLinear layer.}
\label{fig::sparse_grads_is_equiv1}
\end{figure*}

\section{Method}

Our aim is to reduce the amount of trainable parameters at the fine-tuning stage. Taking into account that fine-tuning data is restricted to a limited scope, we assume there is a basis where the weight gradient matrix is very close to being sparse. To identify this basis, we applied a decomposition technique to the stacked weight gradient matrices. As a result, we introduce a new PyTorch layer class, \textbf{SparseGradLinear}, which transitions weights to this sparse gradient space, accumulates gradients in sparse form, and enables the reverse transition back to the original space.

\begin{figure}[h]
\centering
\scalebox{0.95}{
\minipage{0.247\textwidth}

\includegraphics[width=\linewidth]{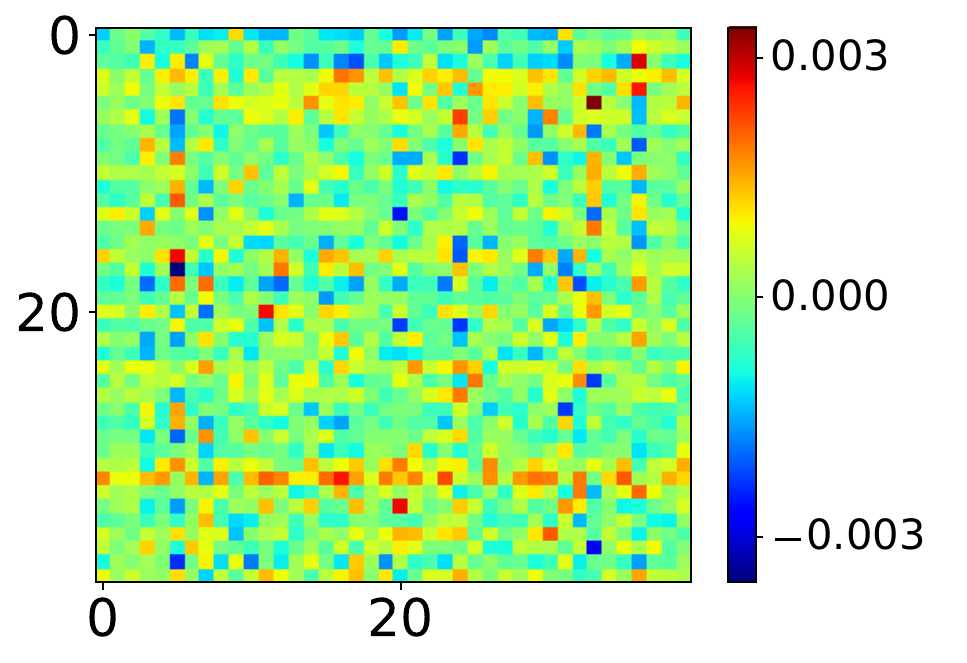}
 \endminipage\hfill
 \minipage{0.24\textwidth}

\includegraphics[width=\linewidth]{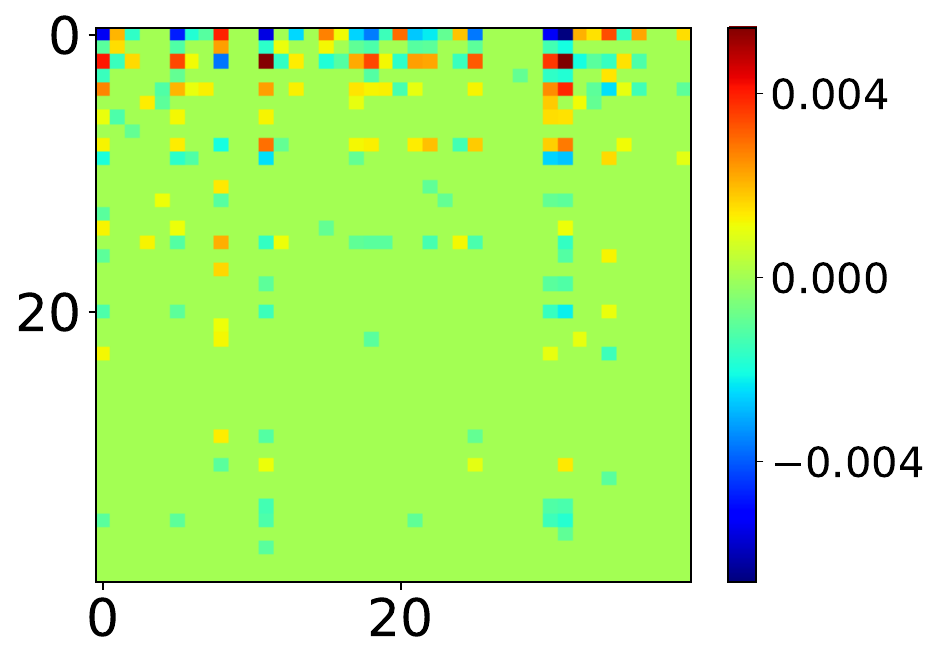}
\endminipage
}
\hspace*{-0.25cm}
\scalebox{0.9}{
\hspace{-0.45cm}
\minipage{0.22\textwidth}
\includegraphics[width=\linewidth]{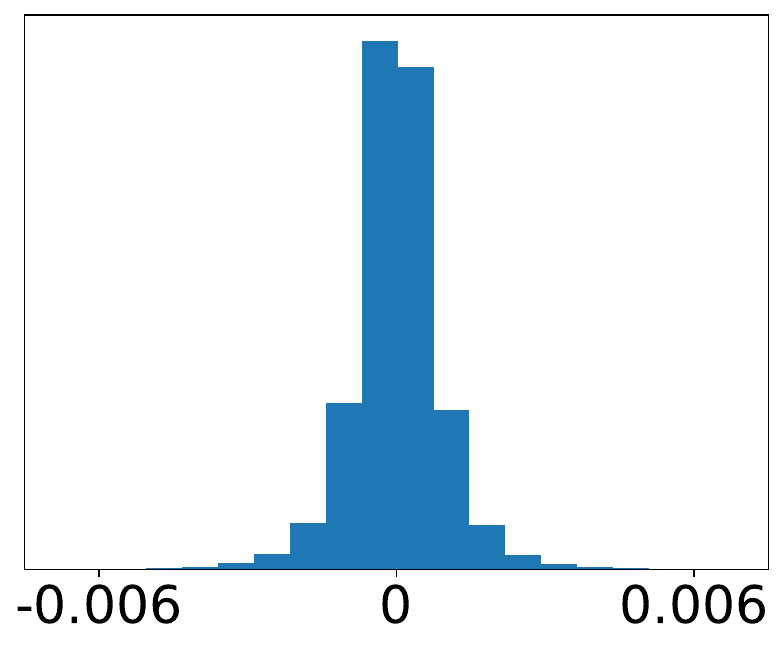}
 \endminipage\hspace{0.15cm}
 \minipage{0.22\textwidth}
\includegraphics[width=\linewidth]{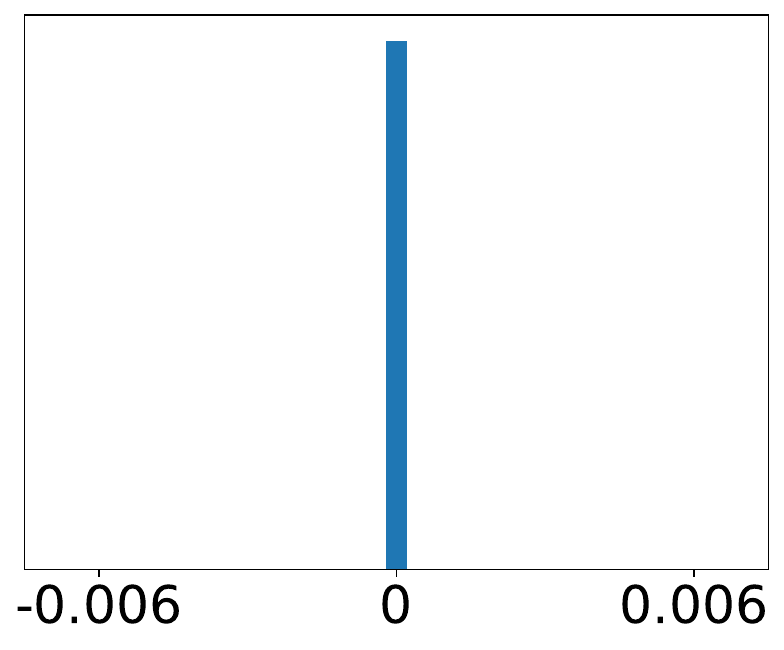}
\endminipage
}
\caption{Gradients on the 5-th BERT MLP: $U \frac{\partial{L}}{\partial{W^T}} V^T$ (right) is more sparse than the original $\frac{\partial{L}}{\partial{W^T}}$ (left).}
\label{fig::sparse_and_dense}
\end{figure}

\subsection{Preliminary Phase: Finding Transition Matrices}

To obtain transition matrices, an initial procedure is necessary. During this, we perform $n\_steps$ of standard backpropagation by freezing the entire model and unfreezing only the linear layers in MLP blocks. We do it to obtain the set of weights gradient matrices $\frac{\partial{L}}{\partial{W}} \in \mathcal{R}^{D\_in \times D\_out}$. Stacking these matrices over $n\_blocks$ -- the number of all blocks in the model -- and over $n\_steps$, we obtain a 3D tensor of size $D\_in \times D\_out \times (n\_steps*n\_blocks)$.

Applying Higher Order SVD~(HOSVD)~\cite{Cichocki_2016} to this tensor yields matrices $U\in \mathcal{R}^{D\_in \times D\_in}$, corresponding to the dimension $D\_in$ and $V^T\in \mathcal{R}^{D\_out \times D\_out}$, corresponding to $D\_out$. In this way, we get two orthogonal transition matrices $U, V^T$ which are shared across all blocks of the model. Multiplying the layer's weight matrix on the left by $U$ and on the right by $V^T$ transforms it into a new space. In this transformed space, the gradient matrix  exhibits greater sparsity compared to the original space. Examples of $\frac{\partial{L}}{\partial{W^T}}$ with and without transition to the new space are shown in Fig.~\ref{fig::sparse_and_dense}.


\subsection{Signal Propagation in SparseGradLinear Layer}

Given a Transformer Linear layer with a weight matrix $W^T$, input activation $X$, and output $Y = XW^T$, we define the gradients of the output, input, and weights as $\frac{\partial{L}}{\partial{Y}}$, $\frac{\partial{L}}{\partial{X}}$, and $\frac{\partial{L}}{\partial{W^T}}$, respectively. To create the corresponding SparseGradLinear layer, we represent the weights in the $U, V^T$ basis, such that the new weights are $\tilde{W}^T = U W^T V^T$. Since the modules following \textbf{SparseGradLinear} remain unchanged in both forward and backward passes, it is crucial to maintain consistency between outputs of the Original Linear Layer $Y$ and the SparseGradLinear layer $\tilde{Y}$, as well as their input gradients $\frac{\partial{L}}{\partial{X}}$ and $\frac{\partial{L}}{\partial{\tilde{X}}}$.

Table~\ref{tab::forward_backward} outlines these adjustments and illustrates the correspondence of variables in Torch Autograd for Linear and SparseGrad layers.

\begin{table}[h]
\caption{Correspondence of variables in Torch Autograd for a regular Linear layer and SparseGradLinear.}
\centering
\scalebox{0.7}{

\begin{tabular}{l| c | c }
\toprule
Variable / Layer &  Linear & SparseGrad  \\
\midrule
Weights & $W^T$ & $\tilde{W}^T = U W^T V^T$ \\
\midrule
Input & $X$ & $\tilde{X} = XU^T$\\
\midrule
Output & $Y = XW^T$ & $\tilde{Y} = XU^T \tilde{W}^T V$ \\
\midrule
Grad Output & $\frac{\partial{L}}{\partial{Y}}$ &  $\frac{\partial{L}}{\partial{Y}} V^T$ \\
\midrule
Grad Input & $\frac{\partial{L}}{\partial{X}} = \frac{\partial{L}}{\partial{Y}} W^T$ &  $\frac{\partial{L}}{\partial{\tilde{X}}} = {\frac{\partial{L}}{\partial{Y}}} U^T \tilde{W}^T V$ \\
\midrule
Grad Weights & $\frac{\partial{L}}{\partial{W}} = {\frac{\partial{L}}{\partial{Y}}}^T X $ &   $\frac{\partial{L}}{\partial{\tilde{W}}} =  V \frac{\partial{L}}{\partial{Y}}^T X U^T $\\
\bottomrule
\end{tabular}
}
\label{tab::forward_backward}
\end{table}

Thus, SparseGradLinear is equivalent to 3 linear layers: first with frozen weights $U^T$, defined by the HOSVD,  second with trainable new weights $\tilde{W}^T = U W^T V^T$, third with frozen weights $V$, defined by the HOSVD. 
A Fig.~\ref{fig::sparse_grads_is_equiv1} shows the propagation of the signal in this structure.


\subsection{Sparse-by-Dense Matrix Multiplication}

We provide the \textbf{SparseGradLinear} class with updated Forward and Backward procedures. However, the addition of multiplications by~$U, V$ into them increased the execution time and affected peak memory in the training loop.

\begin{figure}[h]
\centering
\scalebox{0.98}{
\hspace{-0.2cm}
\minipage{0.26\textwidth}

\includegraphics[width=\linewidth]{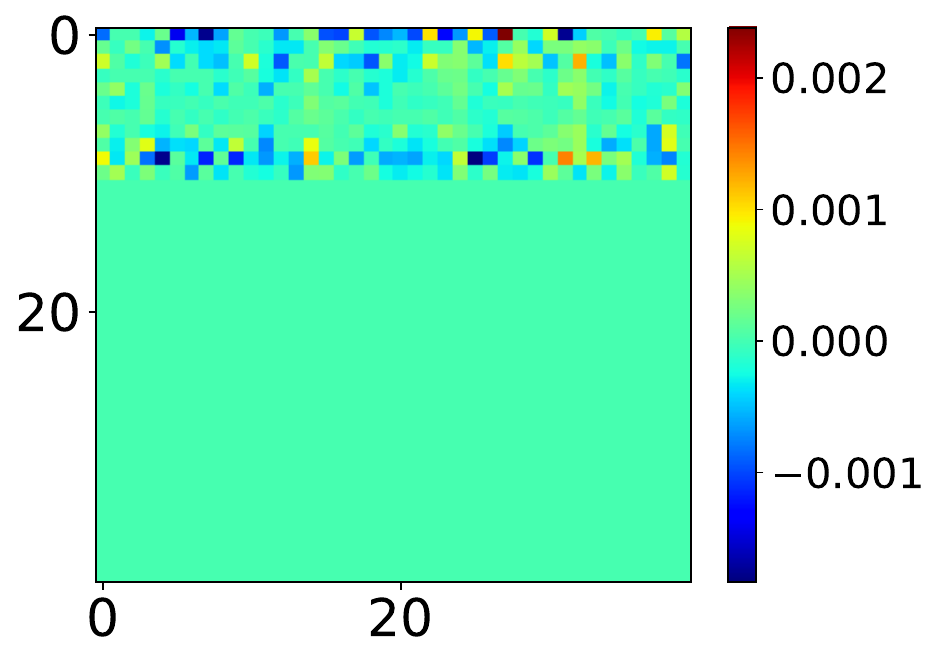}
 \endminipage\hspace{-0.2cm}
 \minipage{0.23\textwidth}

\includegraphics[width=\linewidth]{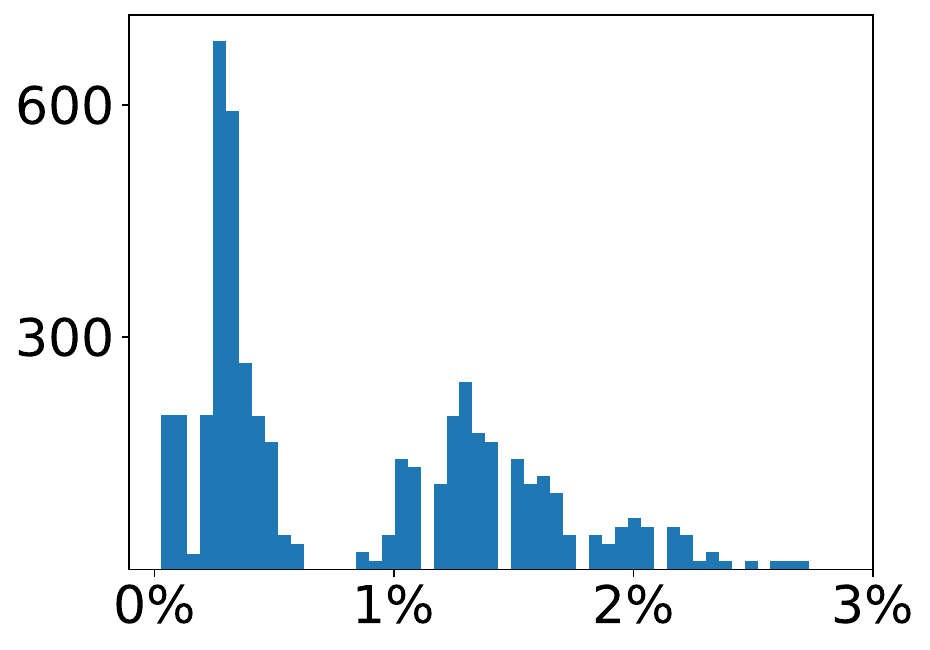}
\endminipage
}

\caption{Strided structure of $\frac{\partial{L}}{\partial{\tilde{Y}}}$~(left) and visualizations of $\%$ nonzero elements in $\frac{\partial{L}}{\partial{\tilde{Y}}}$ throughout training~(right).} 
\label{fig::hist_out_grad}
\end{figure}

The sparsity of the gradient tensor $\frac{\partial{L}}{\partial{\tilde{W}}} = {\frac{\partial{L}}{\partial{\tilde{Y}}}}^TX $ results in some of the multiplicators being sparse.
We explore the structure of each component in this formula and figure out that $\frac{\partial{L}}{\partial{\tilde{Y}}}$ has a sparsity approximately equal to $\frac{\partial{L}}{\partial{\tilde{W}}}$. Histograms of the percent of its non-zero elements are presented in Fig.~\ref{fig::hist_out_grad}. It also shows that the sparsity is "strided" - most of the rows are completely filled with zeros. These rows can be excluded from the multiplication procedure, thus optimizing it.

More precisely, to multiply the sparse matrix $A\in \mathcal{R}^{b \times c}$ by a dense matrix $B\in \mathcal{R}^{c \times d}$ we select $rows$ and $cols$ - indices of rows and columns of $A$ which contain nonzero elements and multiply as follows:
\begin{equation}
C = A(rows,:)(:, cols) B(cols, :).
\end{equation}
We employ $C$ either for further multiplications, or convert it into COO format and send it to SparseAdam optimizer. Indexes in COO format are defined by restoring indexes of $A$: 
\begin{equation}
C_{coo}(rows(k), cols(l)) = C(k, l).
\end{equation}
As it is shown in the Table~\ref{table_ROBERTA_memory_avg}, such procedure significantly speeds up the harnessing of  \textbf{SparseGradLinear}.

\section{Time and Memory Consumption per Training Iteration}
\label{sec:time_memory}

We measure the peak memory allocated during training using the CUDA memory allocator statistics. Table~\ref{table_ROBERTA_memory_avg} demonstrates this statistic on average for all GLUE datasets for the RoBERTa$_{base}$  model.
The comprehensive Tables~\ref{table_ROBERTA_time} and \ref{table_ROBERTA_memory}, which outline metrics for each dataset separately, can be found in Appendix~\ref{sec:appendixB}. Among all methods, LoRA presents the most efficient memory usage, preserving 30\% of the peak memory. SparseGrad, while using slightly more memory, still achieves a 20\% savings. The increase in peak memory with SparseGrad is attributed to the maintenance of matrices $U$ and $V$ and their multiplication by the dense objects, such as Input $X$.

\begin{table}[h]
\fontsize{9.5pt}{9.5pt}
\footnotesize
\centering
\caption{Training speed and memory requirements averaged on the GLUE benchmark. The last two rows of the Table~\ref{table_ROBERTA_memory_avg} report the results for the SparseGrad method with Sparse-by-Dense~(SD) and Regular~(Reg) matrix multiplication, respectively.} 
\scalebox{0.9}{
\begin{tabular}{l | c | c }
\toprule
Method   & Steps/Sec. &  Memory, MB \\
\midrule
Regular FT  &  4.11 & 1345 \\

LoRA  & 4.7 &  944
 \\

SparseGrad\textsubscript{SD} & 4.3 & 1016 \\ 

SparseGrad\textsubscript{Reg} & 0.9 & 1210 \\ 

\bottomrule
\end{tabular}
}
\label{table_ROBERTA_memory_avg}
\end{table}

In terms of training time, LoRA demonstrates the fastest training, followed by SparseGrad, and then standard fine-tuning. Table~\ref{table_ROBERTA_memory_avg} shows that Sparse-by-Dense multiplication saves approximately 12\% memory, leading to an almost five-fold increase in speed.

\begin{table*}[ht]
 \caption{Comparative results of RoBERTa$_{large}$ for 20-epoch task-specific fine-tuning.}
 \footnotesize
 \centering
 \scalebox{0.9}{
 \begin{tabular}{l | l| l| c | c | c | c | c | c | c | c | c }
 \toprule
 Method & \multicolumn{2}{c|}{\#Trainable params} & AVG &  STSB & CoLA & MNLI & MRPC & QNLI & QQP & RTE & SST2 \\
    & Model & MLP block & &  &  &  &  &  &  &  & \\
 \midrule
 Regular FT & 355 mln & 4 mln. & \textbf{85.6} & 91.9\scalebox{0.7}{$\pm .4$} & \textbf{67.1\scalebox{0.7}{$\pm 2.3$}} & \textbf{90.8\scalebox{0.7}{$\pm .2$}} & 89.9\scalebox{0.7}{$\pm .3$} & 92.9\scalebox{0.7}{$\pm .9$} & \textbf{92.3\scalebox{0.7}{$\pm .1$}} & 63.9\scalebox{0.7}{$\pm 7.6$} & 96.7\scalebox{0.7}{$\pm .3$} \\
 \midrule
  LoRA & 168 mln.& 0.05 mln & 83.7 & 92.1\scalebox{0.7}{$\pm .3$} & 64.4\scalebox{0.7}{$\pm .8$} & 90.7\scalebox{0.7}{$\pm .2$} & 89.9\scalebox{0.7}{$\pm .3$} & 93.2\scalebox{0.7}{$\pm .3$} & 91.8\scalebox{0.7}{$\pm .2$} & 60.2\scalebox{0.7}{$\pm 4.1$} & 96.6\scalebox{0.7}{$\pm .1$}
   \\
 SparseGrad & 168 mln. & 0.05 mln & 85.4 & \textbf{92.4\scalebox{0.7}{$\pm .2$}} & 63.2\scalebox{0.7}{$\pm 3.4$} & 90.7\scalebox{0.7}{$\pm .2$} & \textbf{90.5\scalebox{0.7}{$\pm .5$}} & \textbf{93.3\scalebox{0.7}{$\pm .5$}} & 91.7\scalebox{0.7}{$\pm .1$} & \textbf{64.7\scalebox{0.7}{$\pm 6.1$}} & \textbf{96.8\scalebox{0.7}{$\pm .2$}} \\ 
  MeProp & 168 mln. & 0.05 mln & 84.3 & 92.3\scalebox{0.7}{$\pm .1$} & 63.7\scalebox{0.7}{$\pm 1.1$} & 90.4\scalebox{0.7}{$\pm .2$} & 89.4\scalebox{0.7}{$\pm .9$} & 92.5\scalebox{0.7}{$\pm .5$} & 91.4\scalebox{0.7}{$\pm .1$} & 59.2\scalebox{0.7}{$\pm 7.4$} & 96.2\scalebox{0.7}{$\pm .5$} \\

 \bottomrule
 \end{tabular}}
 \label{table_ROBERTA_LARGE_NLU}
 \end{table*}

\section{Experiments}

We conducted experiments on three transformer-based encoder models, BERT and RoBERTa~ \textit{base} and \textit{large}, on the GLUE~\cite{wang2018glue} benchmark, and the LLaMa-2 decoder model on the OpenAssistant Conversations corpus~\cite{kopf2023openassistant}. We compared the fine-tuning of the full model~(Regular FT scheme) with three PEFT methods, namely LoRA, MeProp and SparseGrad, applyed to MLP blocks. To harness LoRA, we use an official repository code. For the MeProp method, we kept the largest elements in the $\frac{\partial L}{\partial W}$ matrix. The proposed SparseGrad involves replacing layers in MLP blocks with its \textbf{SparseGradLinear} equivalents.

\subsection{Natural Language Understanding with BERT and RoBERTa}

We explore the acceptable sparsity level of the gradient matrices in the ``sparse'' space, $\frac{\partial{L}}{\partial{\tilde{W}}}$. By varying the number of remaining parameters in the Linear Layer from $100\cdot 10^3$ to $18\cdot 10^3$, we fine-tuned the model on the GLUE benchmark and identified the point at which performance begins to degrade. This occurs when the number of trainable parameters reaches $22 \times 10^3$, corresponding to $1\%$ of the total weights. Full experimental results can be found in Appendix~\ref{sec:appendixD}.



Guided by this heuristic, in our experiments we leave the top $1\%$ of the largest elements and set the rest to zero. To deal with SparseGradients, we use the SparseAdam optimizer - the masked version of the Adam algorithm. The remaining model parameters are trained with the standard AdamW optimizer.


We fine-tune BERT, RoBERTa$_{base}$ and RoBERTa$_{large}$~\cite{liu2019roberta} using Regular FT, LoRA, MeProp and SparseGrad schemes for $20$ epochs with early stopping for each task in the GLUE. We varied the batch size and learning rate using the Optuna framework~\cite{akiba2019optuna}. The learning rate ranged from $1\mathrm{e}^{-6}$ to $1{\rm e}^{-1}$, and the batch size is selected from the set \{8, 16, 32\}. 
Optimal training parameters for each task are available in the Appendix~\ref{sec:appendixE}. In LoRA we take the rank $10$ for RoBERTa$_{large}$ and rank $7$ for BERT and RoBERTa$_{base}$. For SparseGrad and MeProp we keep the same number of parameters - approximately 1\% of each Linear layer.

The average scores for all GLUE tasks for BERT and RoBERTa$_{base}$ are in the Table~\ref{Tab::avgGlue}; per-task results are placed in the Appendix~\ref{sec:appendixC}.
Table~\ref{table_ROBERTA_LARGE_NLU} depicts the scores for the RoBERTa$_{large}$ model.
Our results indicate that SparseGrad outperforms LoRA with an equivalent number of trainable parameters across all models. For BERT, SparseGrad even exceeds the performance of Regular FT. This may be attributed to the changing basis of the weights in SparseGrad acting as a form of regularization. 
Concerning MeProp, it provides weaker results than SparseGrad in all cases except the RoBERTa$_{large}$ on CoLA. This could be explained by the fact that our approach first transforms the elements into a special ``sparse'' space, while MeProp operates on gradients in the original space. In the original space, the histogram of elements is flatter (see Fig.~\ref{fig::sparse_and_dense}), which suggests that, with the same cut-off threshold, MeProp may remove more significant elements compared to SparseGrad.


\begin{table}[h]
\caption{Average scores over the GLUE benchmark for BERT and RoBERTa\textsubscript{base} models.}
\centering
\scalebox{0.97}{
\footnotesize
\begin{tabular}{l | c| c |c|c  }
\toprule
Model & \multicolumn{2}{c|}{BERT} & \multicolumn{2}{c}{RoB\textsubscript{base}} \\ 
\midrule
Regular FT & 109 mln & 82.5  &  125 mln & \textbf{84.2}\\
LoRA & 54 mln & 81.6 & 68 mln  & 83.1  \\
SparseGrad & 54 mln & \textbf{82.6} & 68 mln  & 83.6 \\
MeProp & 54 mln & 82.1 & 68 mln  & 82.5 \\

\bottomrule
\end{tabular}
}
\label{Tab::avgGlue}
\end{table}

\subsection{Conversations with LLaMa-2}
We apply the SparseGrad method to fine-tune LLaMa-2 7B~\cite{touvron2023llama} model on the OpenAssistant conversational dataset~\cite{kopf2023openassistant}.  
 Fine-tuning was performed on a single GPU NVIDIA A40 during 1 epoch with learning rate $9{\rm e}^{-4}$. For Regular FT, we unfroze \emph{up\_proj} and \emph{down\_proj} layers in the MLP modules with a block index divisible by 3~($0, 3, 6, \dots$). We apply LoRA with rank 32 to the selected blocks, leaving the rest of the model untrainable. 
In the SparseGrad and MeProp methods, we also consider selected MLP modules in the transformer and leave $\approx100,000$~(0,2\%) nonzero elements in the gradient matrix. For LLaMA-2, we conducted a similar ablation study as we did for BERT and RoBERTa. We varied the number of remaining parameters in the MLP block and identified the point where the model's performance began to decline. 

We validate obtained models on the question set MT-Bench Inf from Inflection-Benchmarks~\cite{zheng2023judging}. We followed the guidelines outlined in this work, called "Single Protocol" or "Single Answer Grading”. We got the answers by using the FastChat platform\footnote{\url{https://github.com/lm-sys/FastChat}} and then evaluating them using GPT-4. GPT-4 rates the answers on a scale of 1 to 10, with the evaluation prompt taken from~\cite{zheng2023judging}.

The resulting losses and average GPT-4 scores are presented in Table~\ref{table_LLAMA_opeassist}. While the models perform similarly overall, SparseGrad slightly outperforms LoRA, MeProp, and regular fine-tuning. Examples of responses to Inflection-Benchmark samples are provided in Appendix~\ref{sec:appendixF}. These examples illustrate that, although all models produce good answers, the LoRA-trained model occasionally overlooks important nuances. In the examples given, it fails to recognize that presentations can be stressful for introverts or that hierarchy plays a significant role in Japanese corporate culture.


\begin{table}[h]
\caption{Comparative results for LLaMa-2 on the OpenAssistant-1 dataset.}
\centering
\footnotesize
\scalebox{0.99}{
\begin{tabular}{l | l| c | c}
\toprule
Method & \#Train  &  Valid & I-Bench \\ 
 & params   & Loss & Score \\
 \midrule
Regular FT & 22\% &  1.250 \scalebox{0.7}{$\pm 0.03$} & 4.407  \\

LoRA & 0.5\% & 1.249 \scalebox{0.7}{$\pm 0.05$} & 5.025\\
SparseGrad & 0.5\%  &  1.247 \scalebox{0.7}{$\pm 0.03$} & 5.132\\ 
MeProp & 0.5\%  &  1.259 \scalebox{0.7}{$\pm 0.04$} & 4.261 \\

\bottomrule
\end{tabular}}
\label{table_LLAMA_opeassist}
\end{table}

\section{Conclusion}
We propose a new selective PEFT method called SparseGrad, which identifies a space where the gradients exhibit a sparse structure and updates only its significant part.
SparseGrad is validated through experiments conducted on the BERT, RoBERTa and LLaMa-2 model models, demonstrating its superiority over the additive LoRA and selective MeProp methods. 

Leveraging the sparsity property significantly accelerated the calculations in SparseGrad. Our method runs faster than standard fine-tuning but slower than LoRA, while yielding better performance than LoRA; the same trend applies to memory usage. 
In summary, our method serves as an alternative to LoRA in situations where the performance of the final model takes precedence over the execution time.
The source code as well as links to pretrained models are available at repository\footnote{\url{https://github.com/sayankotor/sparse_grads}}



\section{Limitations}
The main limitation of our method is the additional memory requirements during the Preliminary Phase.
The extra memory is assessed as follows: we need to unfreeze the MLP layers, which hold approximately half of the training parameters in Transformers (see Table~\ref{Tab::parameters_per_layers}), store and decompose a large tensor. For instance, 30 steps in the preliminary phase result in a tensor of approximately 276 MB for BERT and ROBERTA models, and 5.2 GB for LLaMa-2.7 B models. The decomposition part can be the most memory-consuming, as it involves reshaping a 3-dimensional tensor into a matrix with a dimension size equal to the product of two dimension sizes of the tensor~\cite{Cichocki_2016}.

However, this part is executed only once during the entire fine-tuning process and can be computed on the CPU in a short time. The Higher Order SVD decomposition of such objects takes approximately 78 seconds for BERT and RoBERTa$_{base}$ layers and about 668 seconds for LLaMa on an Intel Xeon Gold 6342 CPU processor.

\section{Ethics Statement}
Our proposed approach involves a novel method for fine-tuning large language models, which can be considered as cost-effective as we only update $0.1\%$ of the weights. This type of fine-tuning is environmentally friendly as it reduces resource wastage. We utilized pre-trained models from the Hugging Face repository and implemented updates using the Pytorch library.  We exclusively used open-source datasets to avoid any potential harm or ethical concerns. By prioritizing ethical standards and recognizing potential risks, we strive to promote responsible and sustainable research practices.






\bibliography{acl_latex}

\newpage
\appendix
\onecolumn


\section{Appendix A}
\label{sec:appendixB}
\begin{table*}[h]
\footnotesize
\centering
\scalebox{0.97}{
\begin{tabular}{l | c | c | c | c | c | c | c | c | c }
\toprule
Method / Dataset  & AVG &  STSB & CoLA & MNLI & MRPC & QNLI & QQP & RTE & SST2  \\
 &  &  &  &  &  &  &  &  &  \\

\midrule
Regular FT  &  4.11 & 2.9 &	4.3	& 4.2 &4.1 &3.1	& 4.7 &4.2 & 5.1  \\
\midrule
LoRA  & 4.7 & 2.8	& 5.8 & 6.2 &6.3 & 3.4 & 4.1 &3.2 &4.4 
 \\
SparseGrad, Sparse-by-Dense & 4.3 & 3.8 &1.8 & 3.9 &3.1 &3.5 & 5.6 &6.3 &6.2  \\ 
SparseGrad, Regular & 0.9 & 0.4& 0.3& 0.4& 1.9& 0.8& 0.7& 1.6& 1.1\\

\bottomrule
\end{tabular}}
\caption{The training step execution speed, measured in steps per second (where a higher value indicates faster execution), is reported for the RoBERTa base model. The last two rows describe the SparseGradMethod with Sparse-by-Dense multiplication and with Regular matrix multiplication.}
\label{table_ROBERTA_time}
\end{table*}

\begin{table*}[h]
\footnotesize
\centering
\scalebox{0.97}{
\begin{tabular}{l | l| c | c | c | c | c | c | c | c | c | c }
\toprule
Method / Dataset  & AVG &  STSB & CoLA & MNLI & MRPC & QNLI & QQP & RTE & SST2 \\
 &  &  &  &  &  &  &  &  & \\

\midrule
Regular FT  &  1345 & 1344 & 1358 &1350 & 1362 &1369 &1333 &1314 &1339 \\
\midrule
LoRA  & 944 & 969 & 978	& 986 &998 &938 &935 &902 &855 
 \\
SparseGrad, Sparse-by-Dense & 1016 & 997 & 1082&  1017 & 1110 &1019 & 981 & 960 &980  \\ 
SparseGrad, Regular & 1210 &  1283 & 1212&  1256 & 1183 & 1245 & 1172 & 1116 &1209  \\

\bottomrule
\end{tabular}}
\caption{Peak memory measurement in MB for training loop for the model RoBERTa base.}
\label{table_ROBERTA_memory}
\end{table*}

\onecolumn
\section{Appendix B}
\label{sec:appendixC}

\begin{table*}[h]
\caption{Comparative results of BERT model for 20-epoch task-specific fine-tuning.}
\footnotesize
\centering
\scalebox{0.97}{
\begin{tabular}{l | l |l | c | c | c | c | c | c | c | c | c  }
\toprule
Method & \multicolumn{2}{c|}{\#Trainable}  & AVG &  STSB & CoLA & MNLI & MRPC & QNLI & QQP & RTE & SST2  \\
  & \multicolumn{2}{c|}{Parameters} &  &  &  &  &  &  &  &  &  \\
  & Model & MLP Layer & &  &  &  &  &  &  &  &  \\  
\midrule
Regular FT & 109 mln & 3 mln & 82.5 & \textbf{89.3}\scalebox{0.5}{$\pm .6$} & \textbf{59.0}\scalebox{0.5}{$\pm 1.9$} & 84.0\scalebox{0.5}{$\pm .3$} & 86.2\scalebox{0.5}{$\pm 1.1$} & 89.3\scalebox{0.5}{$\pm 1.3$} & \textbf{91.1}\scalebox{0.5}{$\pm 0$} & 67.4\scalebox{0.5}{$\pm 2.8$} & \textbf{92.7\scalebox{0.5}{$\pm .1$}} \\
\midrule
 LoRA & 53 mln & 0.03 mln& 81.6 & 89.2\scalebox{0.5}{$\pm .7$} & 58.4\scalebox{0.5}{$\pm 2.3$} & \textbf{84.2}\scalebox{0.5}{$\pm .2$} & 83.8\scalebox{0.5}{$\pm .6$} & 89.3\scalebox{0.5}{$\pm .8$} & 91.0\scalebox{0.5}{$\pm 0$} & 64.6\scalebox{0.5}{$\pm 2.1$} & 92.3\scalebox{0.5}{$\pm .2$} \\
  SparseGrad & 53 mln & 0.03 mln & \textbf{82.6} & 89.2\scalebox{0.5}{$\pm .4$} & 58.8\scalebox{0.5}{$\pm 0$} & 84.0\scalebox{0.5}{$\pm 1.3$} & \textbf{86.6}\scalebox{0.5}{$\pm .5$} & \textbf{89.4}\scalebox{0.5}{$\pm 1.6$} & 90.9\scalebox{0.5}{$\pm .3$} & \textbf{69.3\scalebox{0.5}{$\pm 2.9$}} & 92.4\scalebox{0.5}{$\pm .1$}  \\ 
 MeProp & 53 mln & 0.03 mln& 82.1 & 88.9\scalebox{0.5}{$\pm .5$} & 58.4\scalebox{0.5}{$\pm .8$} & 83.3\scalebox{0.5}{$\pm .3$} & 84.2\scalebox{0.5}{$\pm .6$} & 89.6\scalebox{0.5}{$\pm .3$} & 90.4\scalebox{0.5}{$\pm .4$} & 64.9\scalebox{0.5}{$\pm .9$} & 92.1\scalebox{0.5}{$\pm .1$} \\

\bottomrule
\end{tabular}}
\label{table_BERT_NLU}
\end{table*}

 \begin{table*}[h]
 \caption{Comparative results of ROBERTA for 20-epoch task-specific fine-tuning.}
 \footnotesize
 \centering
 \scalebox{0.97}{
 \begin{tabular}{l | l|l| c | c | c | c | c | c | c | c | c}
 \toprule
 Method & \multicolumn{2}{c|}{\#Trainable} & AVG &  STSB & CoLA & MNLI & MRPC & QNLI & QQP & RTE & SST2 \\
   & \multicolumn{2}{c|}{parameters} & &  &  &  &  &  &  &  &  \\
  & Model & MLP Layer & &  &  &  &  &  &  &  &  \\  
 \midrule
 Regular FT & 125 mln.& 3 mln. & \textbf{84.2} & 90.4\scalebox{0.5}{$\pm .3$} & 59.7\scalebox{0.5}{$\pm 1.4$} & \textbf{87.7}\scalebox{0.5}{$\pm .1$} & \textbf{90.0}\scalebox{0.5}{$\pm .6$} & \textbf{90.6}\scalebox{0.5}{$\pm .8$} & \textbf{91.5}\scalebox{0.5}{$\pm .1$} & \textbf{68.8\scalebox{0.5}{$\pm 2.5$}} & \textbf{94.7\scalebox{0.5}{$\pm .2$}} \\
 \midrule
  LoRA & 68 mln.& 0.03 mln. & 83.1 & 90.5\scalebox{0.5}{$\pm .2$} & \textbf{60.6}\scalebox{0.5}{$\pm 1.7$} & 87.5\scalebox{0.5}{$\pm .1$} & 88.4\scalebox{0.5}{$\pm .6$} & 90.0\scalebox{0.5}{$\pm .8$} & 91.4\scalebox{0.5}{$\pm .1$} & 63.1\scalebox{0.5}{$\pm 2.3$} & 94.5\scalebox{0.5}{$\pm .1$} 
   \\
 SparseGrad & 68 mln.& 0.03 mln. & 83.6 & \textbf{90.8}\scalebox{0.5}{$\pm .2$} & 60.0\scalebox{0.5}{$\pm 1.6$} & 87.5\scalebox{0.5}{$\pm .1$} & 89.6\scalebox{0.5}{$\pm 1.1$} & 91.5\scalebox{0.5}{$\pm .6$} & 91.5\scalebox{0.5}{$\pm .1$} & 65.6\scalebox{0.5}{$\pm 2.1$} & 94.2\scalebox{0.5}{$\pm .1$}\\ 

 MeProp & 68 mln.& 0.03 mln. & 82.5 & 90.7\scalebox{0.5}{$\pm .1$} & 59.2\scalebox{0.5}{$\pm 1.3$} & 85.9\scalebox{0.5}{$\pm .1$} & 89.1\scalebox{0.5}{$\pm 0.9$} & 89.4\scalebox{0.5}{$\pm .5$} & 90.5\scalebox{0.5}{$\pm .1$} & 61.5\scalebox{0.5}{$\pm 1.6$} & 94.2\scalebox{0.5}{$\pm .1$}\\

 \bottomrule
 \end{tabular}}
 \label{table_ROBERTA_NLU}
 \end{table*}

\onecolumn
\section{Appendix C}
\label{sec:appendixD}
The average GLUE results for the BERT and RoBERTa$_{base}$ models with respect to the number of remaining updated parameters in Linear layers. Tables~\ref{table_BERT_sparsity},~\ref{table_ROBERTA_sparsity} shows that under the 0.8\% of the remaining parameters, performance tends to decrease.


    
    

    

\begin{table*}[ht]
\caption{GLUE score as a function of the weight gradient sparsity in BERT}
\centering
\scalebox{0.77}{
\begin{tabular}{l | c | c | c | c | c | c | c | c | c | c }
\toprule
Method & \% of remained & AVG &  STSB & CoLA & MNLI & MRPC & QNLI & QQP & RTE & SST2  \\
  & params & &  &  &  &  &  &  &  &  \\
  & in Linear Layers & &  &  &  &  &  &  &  &  \\  
\midrule
SparseGrad & 100 & 82.6 & 89.2\scalebox{0.5}{$\pm .4$} & 58.8\scalebox{0.5}{$\pm 0$} & 84.0\scalebox{0.5}{$\pm 1.3$} & 86.6\scalebox{0.5}{$\pm .5$} & 89.4\scalebox{0.5}{$\pm 1.6$} & 90.9\scalebox{0.5}{$\pm .3$} & 69.3\scalebox{0.5}{$\pm 2.9$} & 92.4\scalebox{0.5}{$\pm .1$}  \\ 
\midrule
 SparseGrad 18k & 0.8 & 81.5 & 89.1\scalebox{0.5}{$\pm .3$} & 59.1\scalebox{0.5}{$\pm .5$} & 83.8 \scalebox{0.5}{$\pm .1$}& 84.6\scalebox{0.5}{$\pm .8$} & 89.4\scalebox{0.5}{$\pm .8$} & 90.8\scalebox{0.5}{$\pm .2$} & 63.5\scalebox{0.5}{$\pm 5.2$} & 92.4\scalebox{0.5}{$\pm .6$}  \\ 
SparseGrad 22k & 1 &82.2 & 89.7\scalebox{0.5}{$\pm .6$} & 60.0\scalebox{0.5}{$\pm .4$} & 83.9\scalebox{0.5}{$\pm .1$} & 84.6\scalebox{0.5}{$\pm 1.5$} & 88.8\scalebox{0.5}{$\pm 1.0$} & 91.1\scalebox{0.5}{$\pm .0$} & 67.7\scalebox{0.5}{$\pm 2.5$} & 92.3\scalebox{0.5}{$\pm .3$}  \\ 
 SparseGrad 30k & 1.2 & 82.0 & 89.2\scalebox{0.5}{$\pm .4$} & 59.1\scalebox{0.5}{$\pm .5$} & 84.1\scalebox{0.5}{$\pm .3$} & 85.4\scalebox{0.5}{$\pm .6$} & 89.3\scalebox{0.5}{$\pm .6$} & 90.8\scalebox{0.5}{$\pm .2$} & 65.6\scalebox{0.5}{$\pm 4.6$} & 92.2\scalebox{0.5}{$\pm .4$}  \\ 
 SparseGrad 100k & 4.2 & 82.2 & 89.3\scalebox{0.5}{$\pm .3$} & 60.0\scalebox{0.5}{$\pm .3$}  &  83.8 \scalebox{0.5}{$\pm .2$}& 85.1\scalebox{0.5}{$\pm 1.2$} & 88.9\scalebox{0.5}{$\pm 1.0$} & 91.2\scalebox{0.5}{$\pm .0$} & 65.6\scalebox{0.5}{$\pm 3.3$} & 92.4\scalebox{0.5}{$\pm .3$}  \\ 

\bottomrule
\end{tabular}}
\label{table_BERT_sparsity}
\end{table*}

\begin{table*}[ht]
\caption{GLUE score as a function of the weight gradient sparsity in ROBERTA}
\centering
\scalebox{0.77}{
\begin{tabular}{l | c | c | c | c | c | c | c | c | c | c  }
\toprule
Method  & \% of remained & AVG &  STSB & CoLA & MNLI & MRPC & QNLI & QQP & RTE & SST2  \\
  & params &  &  &  &  &  &  &  &  &  \\
  & in Linear Layers & &  &  &  &  &  &  &  &  \\  
\midrule
 SparseGrad & 100 & 83.6 & 90.8\scalebox{0.5}{$\pm .2$} & 60.0\scalebox{0.5}{$\pm 1.6$} & 87.5\scalebox{0.5}{$\pm .1$} & 89.6\scalebox{0.5}{$\pm 1.1$} & 91.5\scalebox{0.5}{$\pm .6$} & 91.5\scalebox{0.5}{$\pm .1$} & 65.6\scalebox{0.5}{$\pm 2.1$} & 94.2\scalebox{0.5}{$\pm .1$}\\ 
\midrule
 SparseGrad 18k & 0.8 & 83.4 & 90.9\scalebox{0.5}{$\pm .2$} & 59.7\scalebox{0.5}{$\pm .1$}  & 87.4  \scalebox{0.5}{$\pm .4$}& 89.2\scalebox{0.5}{$\pm .7$} & 89.1\scalebox{0.5}{$\pm .4$} & 91.5\scalebox{0.5}{$\pm .1$} & 60.4\scalebox{0.5}{$\pm 5.8$} & 94.0\scalebox{0.5}{$\pm .4$}  \\ 
 SparseGrad 22k & 1 & 83.6 & 90.6\scalebox{0.5}{$\pm .2$} & 58.8\scalebox{0.5}{$\pm .4$} & 87.7 \scalebox{0.5}{$\pm .1$}& 90.0\scalebox{0.5}{$\pm .3$} & 90.1\scalebox{0.5}{$\pm .1$} & 91.3\scalebox{0.5}{$\pm .1$} & 65.5\scalebox{0.5}{$\pm 3.7$} & 94.6\scalebox{0.5}{$\pm .2$}  \\ 
 SparseGrad 30k & 1.2 & 83.6 & 90.8\scalebox{0.5}{$\pm .3$} & 59.4\scalebox{0.5}{$\pm .4$} & 87.6 \scalebox{0.5}{$\pm .1$}& 89.8\scalebox{0.5}{$\pm .4$} & 91.0\scalebox{0.5}{$\pm .1$} & 91.3\scalebox{0.5}{$\pm .1$} & 64.9\scalebox{0.5}{$\pm 8.6$} & 94.2\scalebox{0.5}{$\pm .1$} \\ 
 SparseGrad 100k &1.4  & 83.9 & 90.9\scalebox{0.5}{$\pm .3$} & 59.8\scalebox{0.5}{$\pm .5$} & 87.0\scalebox{0.5}{$\pm .1$}&89.7 \scalebox{0.5}{$\pm .5$}& 89.6\scalebox{0.5}{$\pm .5$} & 91.4\scalebox{0.5}{$\pm .1$} & 69.4\scalebox{0.5}{$\pm 3.2$} & 94.1\scalebox{0.5}{$\pm .3$} \\ 

\bottomrule
\end{tabular}}
\label{table_ROBERTA_sparsity}
\end{table*}

\onecolumn
\section{Appendix D}
\label{sec:appendixE}
Best training parameters for all models. In all experiments, we repeat fine-tuning $3$ times over different seeds and report the average score.
\begin{table}[h]
\caption{Best training parameters on GLUE benchmark for BERT model.}
\footnotesize
\centering
\scalebox{0.97}{
\begin{tabular}{l | c | c }
\toprule
Dataset &  batch size &  learning rate  \\
 \midrule
  STSB & 32 & 1.24e-4     \\
 \midrule
  CoLA & 32 & 3.15e-5    \\
\midrule
  MNLI & 32 & 6.07e-6     \\
\midrule
 MRPC &  32 & 1.22e-5     \\
\midrule
 QNLI & 16 & 1.94e-5    \\
\midrule
  QQP & 32 & 1.41e-5     \\
\midrule
  RTE & 16 & 6.81e-5     \\
\midrule
  SST2 & 32 & 1.47e-5     \\

\bottomrule
\end{tabular}}
\label{BERT Sparse Parameters}
\end{table}

\begin{table}[h]
\caption{Best training parameters on GLUE benchmark for RoBERTa model.}
\footnotesize
\centering
\scalebox{0.97}{
\begin{tabular}{l | c | c }
\toprule
Dataset &  batch size &  learning rate  \\
 \midrule
  STSB & 16 & 2.70e-5  \\
 \midrule
  CoLA & 16 & 1.01e-5     \\
\midrule
  MNLI & 32 & 1.51e-5     \\
\midrule
 MRPC & 32 & 1.9e-5     \\
\midrule
 QNLI & 16 & 1.91e-5     \\
\midrule
  QQP & 16 & 5.11e-6    \\
\midrule
  RTE & 32 & 3.05e-5     \\
\midrule
  SST2 & 16 & 1.33e-5    \\

\bottomrule
\end{tabular}}
\label{ROBERTA Sparse Parameters}
\end{table}

\begin{table}[h]
\caption{Best training parameters on GLUE benchmark for RoBERTa-large model.}
\footnotesize
\centering
\scalebox{0.97}{
\begin{tabular}{l | c | c }
\toprule
Dataset &  batch size &  learning rate  \\
 \midrule
  STSB & 32 & 7.71e-5    \\
 \midrule
  CoLA & 16 & 1.8e-5     \\
\midrule
  MNLI & 16 & 1.15e-6     \\
\midrule
 MRPC & 32 & 2.47e-5     \\
\midrule
 QNLI & 16 & 8.83e-6     \\
\midrule
  QQP & 32 & 7.2e-6     \\
\midrule
  RTE & 32 & 1.02e-5     \\
\midrule
  SST2 & 32 & 1.02e-5     \\

\bottomrule
\end{tabular}}
\label{ROBERTA Large Sparse Parameters}
\end{table}

\onecolumn
\section{Appendix E}
\label{sec:appendixF}
Responses from the models to an example from Inflection-Benchmarks are shown. While all models perform fairly well, the LoRA-trained model overlooks the fact that public speaking can be stressful for an introvert when answering the first question.
\begin{tcolorbox}[colframe=black, colback=white, boxrule=0.4mm, arc=2mm]
\footnotesize
\textbf{[Question 1]}\\
Write a persuasive email to convince your introverted friend, who dislikes public speaking, to volunteer as a guest speaker at a local event. Use compelling arguments and address potential objections. Please be concise.

\vspace{1em}
\textbf{LLaMa2-7B Regular FT:}\\

Here's a persuasive email to convince your introverted friend to volunteer as a guest speaker at a local event:

Subject: Join us as a guest speaker at our local event

Dear (Friend's Name),
I hope this email finds you well. I hope you're doing well! I'm writing to invite you to be a guest speaker at our upcoming local event. As a friend, I know how passionate you are!

\vspace{1em}
\textbf{LLaMa2-7B LoRA:}\\

Subject Line: Please, Volunteer as a Guest Speaker!
Dear John,
I hope you are doing well. I am writing to you because I know you are an introverted person and do not like public speaking. However, I have a great opportunity for you to volunteer as a guest speaker at the evant!

\vspace{1em}
\textbf{LLaMa2-7B Sparse Grad:}\\

Subject: Volunteer as a Guest Speaker at a Local Event

Dear [Name],
I hope this email finds you well. I wanted to reach out to you about an exciting opportunity to volunteer as a guest speaker at a local event.
As you know, public speaking can be a nerve-wracking experience, but it's also a great way to share your knowledge and expertise with others.

\vspace{1em}

\textbf{[Question 2]}\\
What are some business etiquette norms when doing business in Japan?

\vspace{1em}
\textbf{LLaMa2-7B Regular FT:}\\

If you're doing business in Japan, it's important to follow their business etiquette norms. Here are some key things to keep in mind:

1. Hierarchy: In Japan, it's important to respect the hierarchy in the workplace. Address your Japanese colleagues with the appropriate title, and don't speak to them unless they initiate the conversation.

2. Punctuality: Being punctual is considered very important in Japanese culture, so make sure to arrive on time for meetings and events.

3. Business Cards: Business cards are an important part of doing business in Japan. Make sure to have your own business card made beforehand and carry it with you at all times.

\vspace{1em}
\textbf{LLaMa2-7B LoRA:}\\

Of course! Here are some business etiquette norms when doing business in Japan:

1. Business Cards: Business cards are a vital part of business interactions in Japan. Always have a few business cards with you and exchange them with others during introductions.

2. Punctuality: Time is sacred in Japan, and being punctual is essential for business meetings. Arrive on time or even a few minutes early.

3. Gifts: Gifts are common in business interactions in Japan, especially for special occasions. Small personalized gifts such as chocolates or flowers are appropriate.

\vspace{1em}
\textbf{LLaMa2-7B Sparse Grad:}\\

Here are some business etiquette norms to keep in mind when doing business in Japan:

1. Respect for seniority: In Japanese culture, age and seniority are highly respected, so it is important to show respect to those who are older or in a higher position than you.

2. Business cards: Business cards are an important part of doing business in Japanese culture, and it is customary to exchange business cards at the beginning of a meeting.

3. Gifts: Gifts are often given as a sign of appreciation or gratitude in Japanese culture. It is important to give gifts that are appropriate for the occasion and not too extravagant.

\end{tcolorbox}

\end{document}